\documentclass[letterpaper, 10 pt, conference]{ieeeconf}  

\IEEEoverridecommandlockouts                              

\usepackage{graphics} 
\usepackage{epsfig} 
\usepackage{mathptmx} 
\usepackage{times} 
\usepackage{amsmath} 
\usepackage{amssymb}  
\usepackage{balance}

\title{\Large \bf Aerial Animal Biometrics: Individual Friesian Cattle Recovery and Visual Identification via an Autonomous UAV with Onboard Deep Inference\vspace{-7pt}}

\author{William Andrew$^{1}$, Colin Greatwood$^{2}$ and Tilo Burghardt$^{1}$
\thanks{*This work was supported by the EPSRC Centre for Doctoral Training in Future Autonomous and Robotic Systems (FARSCOPE) at the Bristol Robotics Laboratory (BRL).}
\thanks{$^{1}$Department of Computer Science, $^{2}$Department of Airspace Engineering, Faculty of Engineering, University of Bristol, United Kingdom (UK).}%
}

\usepackage{color}

\begin{document}

\maketitle

\thispagestyle{empty}
\pagestyle{empty}

\begin{abstract}
This paper describes a computationally-enhanced M100 UAV platform with an onboard deep learning inference system for integrated computer vision and navigation.
The system is able to autonomously find and visually identify by coat pattern individual Holstein Friesian cattle in freely moving herds. 
We propose an approach that utilises three deep convolutional neural network  architectures running live onboard the aircraft: \textit{(1)} a YOLOv2-based  species detector, \textit{(2)} a dual-stream deep network delivering exploratory agency, and \textit{(3)} an InceptionV3-based biometric long-term recurrent convolutional network for individual animal identification. 
We evaluate the performance of each of the components offline, and also online via real-world field tests comprising 147~minutes of autonomous low altitude flight  in a  farm environment  over a dispersed herd of 17~heifer~dairy~cows.
We report error-free identification  performance on this online experiment. 
The presented proof-of-concept  system   is the first  of its kind. It represents a practical     step towards   autonomous biometric  identification of individual   animals from the air in open pasture environments for tag-less AI support in farming and ecology.  
\end{abstract}


\section{Introduction and Related Work}

\begin{figure}[t]
\vspace{5pt}
\includegraphics[width=245pt,height=272pt]{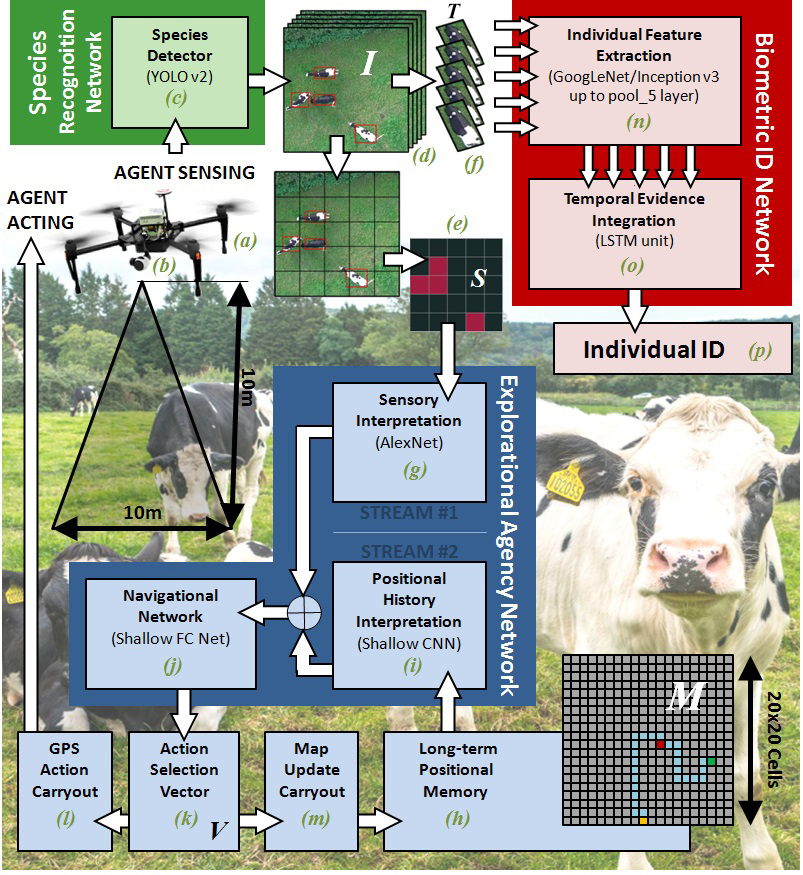}
\caption{\textbf{UAV Onboard System Overview.} For individual Friesian cattle search and identification we propose an approach that utilises three deep convolutional neural architectures operating onboard a \textit{\textbf{(a)}} computationally-enhanced DJI Matrice 100 platform to achieve species detection~(green), exploratory agency~(blue), and individual animal ID~(red). 
In particular, a \textit{\textbf{(b)}} DJI Zenmuse X3 camera stream reshaped to $720\times 720$ images is processed by \textit{\textbf{(c)}} a YOLOv2-based Friesian cattle detector yielding \textit{\textbf{(d)}} RoI-annotated frames~$I$.  These are transformed into an \textbf{\textit{(e)}}~$5\times 5$ occupancy grid map~$S$ and also a set of \textit{\textbf{(f)}}~spatio-temporal tracklets~$\{T_0, T_1, ..., T_p\}$. 
The map~$S$ is input to a \textit{\textbf{(g)}}~tactical network based on AlexNet, which forms a dual stream navigation architecture together with a \textit{\textbf{(i)}} strategic network operating on \textbf{\textit{(h)}}~a long-term exploratory history memory $M$. 
Both streams are concatenated into \textbf{\textit{(j)}} a shallow navigation net that outputs \textit{\textbf{(k)}} a score vector~$V$ based on which possible navigational actions $a\in\{N,W,S,E\}$ are selected. 
During operation selected actions are \textbf{\textit{(l)}} performed grounded in GPS and, in-turn, \textit{\textbf{(m)}} the positional history~$M$ is
updated. 
For individual ID, \textit{\textbf{(f)}} each tracklet $T$ is re-scaled and passed into an \textit{\textbf{(n)}}~Inception~V3~network up to the pool\_5 layer followed by \textit{\textbf{(o)}} a LSTM unit for temporal information integration mapping to \textit{\textbf{(p)}}~a vector over the individual cattle IDs.}
\label{fig:overview}\vspace{-12pt}
\end{figure}

This paper presents an unmanned aerial vehicle~(UAV) platform with onboard deep learning inference~(see Fig.~\ref{fig:overview}) that  autonomously locates and visually identifies individual Holstein Friesian cattle by their uniquely-coloured coats in low altitude flight (approx. 10m) within a geo-fenced farm area. 
The task encompasses the integrated performance of species detection, exploratory agency, and individual animal identification (ID). 
All tasks are performed entirely onboard a custom DJI M100 quadrotor with limited computational resources, battery lifetime, and payload size.

In doing so, this work attempts to assist agricultural monitoring in performing minimally-invasive cattle localisation and identification in the field. Possible applications include the behavioural analysis of social hierarchies~\cite{ungerfeld2014time, kondo1990stabilization, phillips2002effects}, grazing patterns \cite{gregorini2012diurnal, gregorini2006behavior} and herd~welfare~\cite{sowell2000social}.

The search for targets  with unknown locations traditionally arises in search and rescue~(SAR) scenarios \cite{lin2009uav, waharte2010supporting, ryu2012autonomous}. 
In visually-supported navigation for this task, approaches broadly  operate either a map-based or map-less paradigm~\cite{bonin2008visual,desouza2002vision}. 
Map-less approaches have no global environment representation and traditionally operate using template appearance matching~\cite{matsumoto1996visual,jones1997appearance}, optical-flow guidance~\cite{Santos1993divergent}, or landmark feature tracking~\cite{pears2001ground,saeedi2006vision}. 
More recently, such systems have been replaced with  visual input classification via convolutional neural networks~(CNNs)~\cite{giusti2016machine,ran2017convolutional}. 
In this work, we  build on a simulation setup presented in \cite{andrew2018deep} and formulate a 2D global grid approximation of the environment (see map $M$ in Fig.~\ref{fig:overview}) for storing visited positions, current location, and successful target recoveries. 
This concept is inspired by occupancy grid maps~\cite{borenstein1990real,oriolo1995line},
as opposed to post-exploration maps~\cite{moravec1983stanford}
or topological maps~\cite{kosaka1992fast}. 
For our cattle recovery task -- and despite their simplicity -- grid maps
still represent a highly effective tool~\cite{andrew2018deep} for exploring the solution space of AI solutions~\cite{Zhang2017neural,melnikov2014projective}.

\newpage

Coat pattern identification of individual Friesian cattle represents a form of animal biometrics~\cite{kuhl2013animal}. Early systems for the particular task at hand utilised the Scale-Invariant Feature Transform (SIFT)~\cite{lowe1999object} on image sequences~\cite{martinez2013video} or Affine~SIFT (ASIFT)~\cite{morel2009asift} to map  from dorsal cow patterns to animal IDs~\cite{Andrew2016automatic}. However, for up-to-date performance we base our  individual ID component  on recent CNN-grounded biometric work~\cite{andrew2017visual} where temporal stacks of the region of interest (RoIs) around detected cattle are analysed by  a Long-term Recurrent Convolutional Network~(LRCN)~\cite{donahue2015long} as shown in Fig.~\ref{fig:overview} in red.
This architecture represents a compromise between light-weight  onboard operation and  more high-end networks with heavier computational~footprints.

Whilst aerial wildlife census applications  routinely use  manually controlled UAVs~\cite{hodgson2013unmanned, koski2009evaluation, abd2005development, rodriguez2012eye},  and have experimented with part-automated photographic gliders~\cite{sloopflyer}, to the best of our knowledge this paper presents the first proof-of-concept system for \textit{fully autonomous exploration} \textit{and} \textit{online} \textit{individual biometric  identification} of  animals onboard an aircraft. 

To summarise this paper's principal contributions:
\begin{itemize}
  \item Proof-of-concept in the viability of autonomous aerial biometric  cattle ID  in a real-world agricultural setting.
        \item Novel combination of algorithms performing online~target detection, identification and exploratory agency.
        \item Validation of the employed UAV hardware setup capable of deep inference onboard the flight platform itself.
      
        \item Real-world and live application of the exploratory simulation framework developed in our previous work \cite{andrew2018deep}.
\end{itemize}

\begin{figure}[b]
\includegraphics[width=160pt,height=118pt]{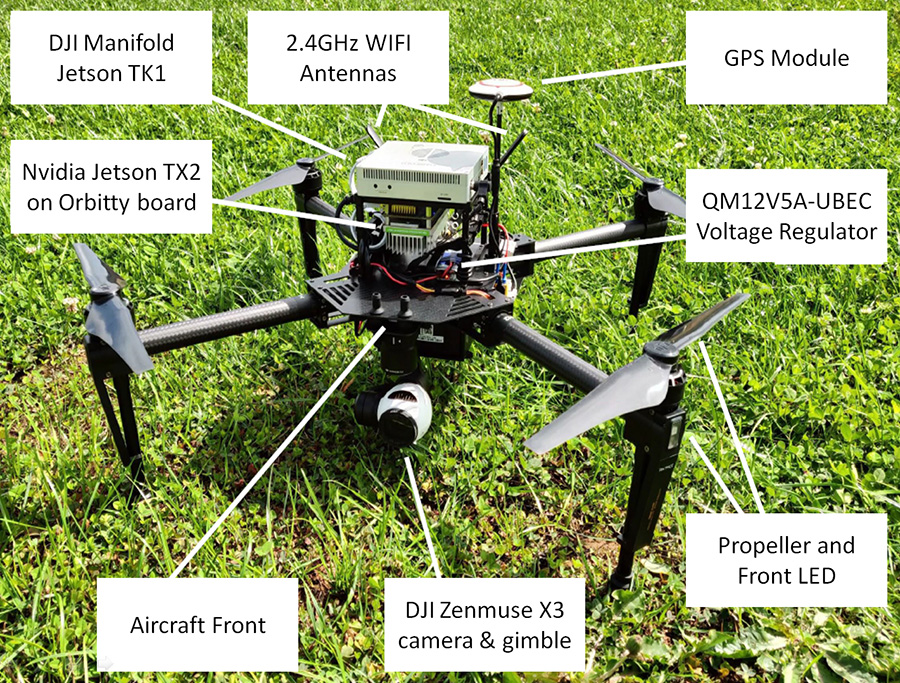}\hspace{1pt}
\includegraphics[width=82pt,height=118pt]{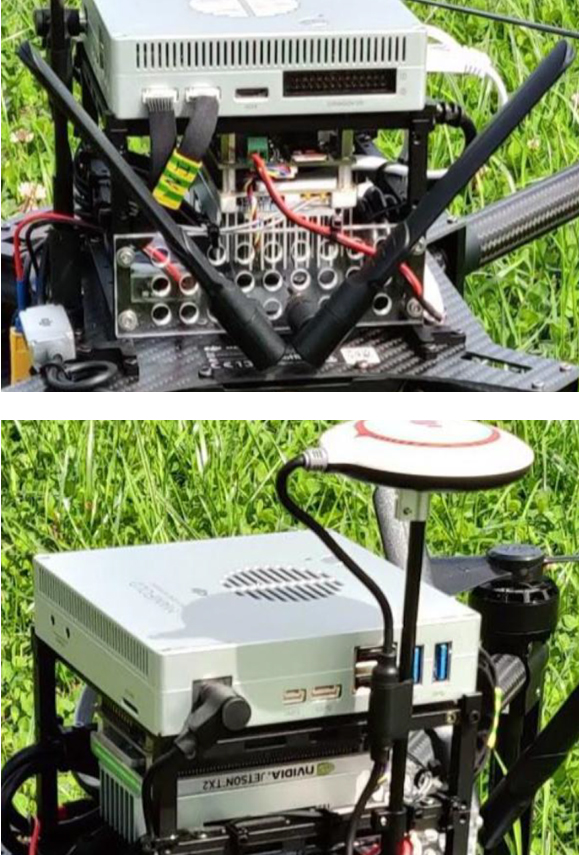}\vspace{-8pt}
\caption{\textbf{Physical UAV Platform.} \textit{\textbf{(left)}} Front view of the customised and fully assembled  DJI Matrice 100 UAV flight platform with selected individual components highlighted.  \textit{\textbf{(right)}} Close-ups of the rear and side of the aircraft revealing custom mounts for two centrally fixed onboard computers, WiFi antennas and the GPS unit.}
\label{fig:uav}
\end{figure}

\section{Hardware}
\begin{figure}[t]
\vspace{5pt}
\includegraphics[width=245pt,height=166pt]{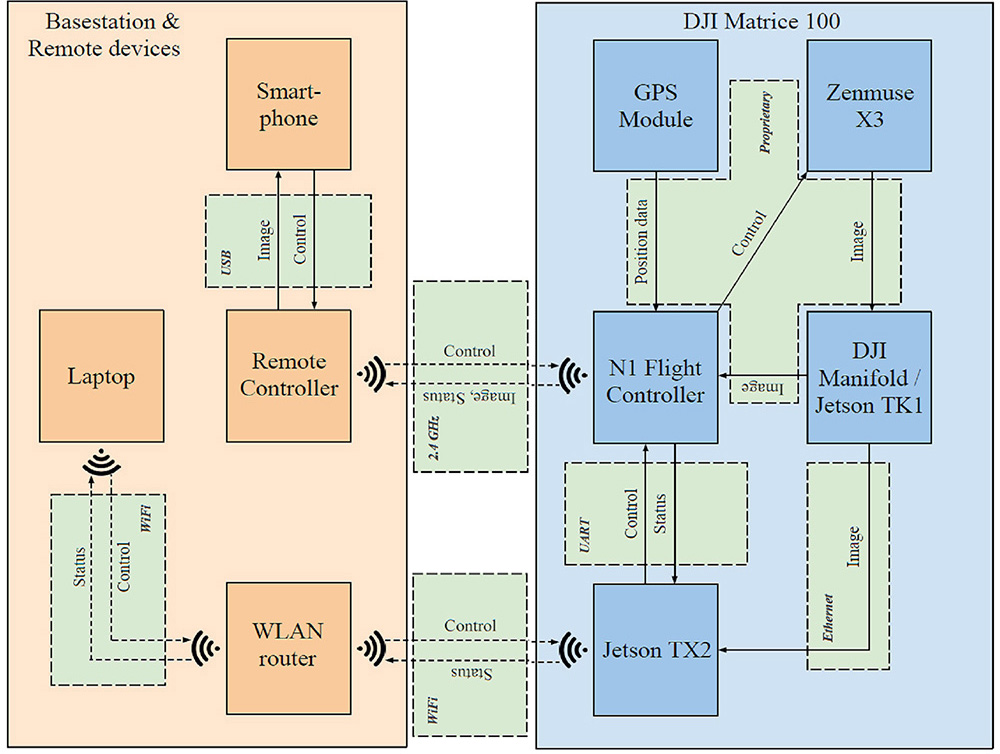}\vspace{-8pt}
\caption{\textbf{Hardware Communication Architecture.} Communication interfaces  (in green) between individual components on the aircraft (in blue) and the base station (in orange). Fully manual aircraft control backup is provided by  the remote system where a live camera feed is visible on an attached smart device. All programmatic commands are issued via ROS-based API calls over a serial connection between the Jetson TX2  and the N1 flight controller. Control of this form is autonomous and is initiated remotely via SSH over a WiFi connection and monitored live via ROS. The DJI Manifold decodes and forwards live imagery from the Zenmuse X3 whilst the Jetson TX2 performs the deep inference via the associated Nvidia GPGPU.}\vspace{-10pt}
\label{fig:coms}
\end{figure}

We use the  DJI Matrice 100 quadrotor UAV, which houses the ROS-enabled DJI N1 flight controller, as our  base platform. 
It has been employed previously across various autonomous tasks \cite{hui2018vision, cobo2016approach, yu2017intelligent, kyristsis2016towards}.
We extend the base M100 platform by adding an Nvidia Jetson TX2 mounted on a Connect Tech Inc. Orbitty carrier board to enable onboard deep inference via 256 CUDA cores under Nvidia's  Pascal\textsuperscript{TM} architecture.
Also onboard is the DJI Manifold~(essentially an Nvidia Jetson TK1) to decode the raw image feed from the onboard camera (the DJI Zenmuse X3 camera/gimbal) and to add further computational capacity.
The X3 camera is mounted on rubber grommets and a 3-axis gimbal, which allows for rotor vibration isolation and independent movement of the flight platform for stabilised footage and programmatically controlled roll-pitch-yaw. 
A  Quanum QM12V5A-UBEC voltage regulation device was fitted to power non-conformal devices feeding off the primary aircraft battery. 
In addition, customised mounts were added for WiFi antennas to monitor the craft remotely. 
Figure~\ref{fig:uav} depicts the complete aircraft with  views of custom components, whilst Figure \ref{fig:coms} shows a detailed overview of the communication infrastructure. 
Note that the base station and remote control devices act in a supervisory role only; all inputs and autonomous control are processed and issued onboard the UAV.

\section{Experimental Setup}
\label{sec:7-full:dataset}

\subsection{ Location, Timing and Target Herd}

\begin{figure}[t]
\vspace{5pt}
 \includegraphics[width=105pt,height=135pt]{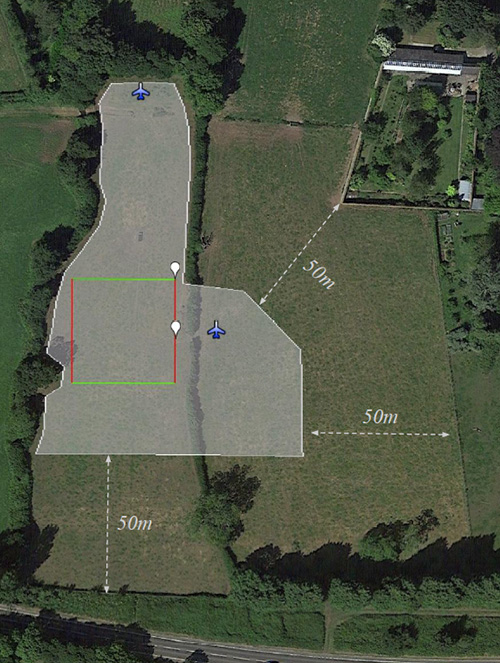}\ \includegraphics[width=138pt,height=135pt]{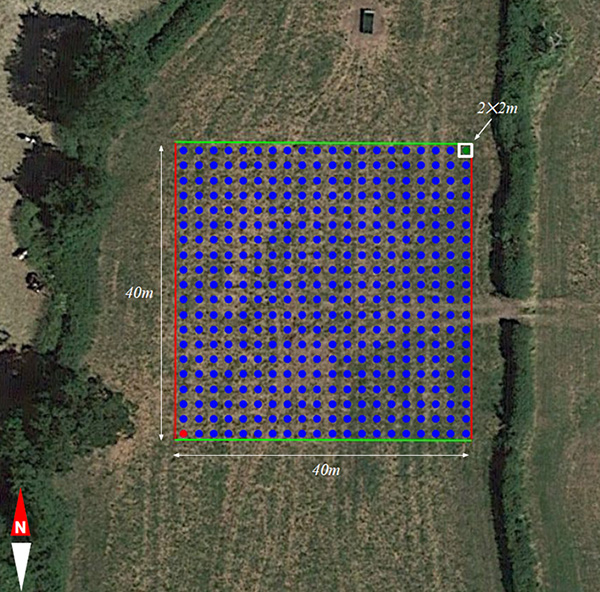}\vspace{-8pt}
\caption{\textbf{Test Environment at Wyndhurst Farm.} Both training data acquisition and autonomous test flights were performed at the University of Bristol's Wyndhurst Farm in Langford Village, UK. \textit{\textbf{(left)}} Illustration of the enforced polygonal geo-fence~(transparent
white area) defined by a set of GPS coordinates and the~$40\times 40$ meters autonomous flight area within, Also shown are two possible take-off and landing sights. \textit{\textbf{(right)}} Autonomous flight area with its $400$ grid cells of $2\times 2$ meters each. White squares show the grid origin along with cell dimensions.}
\label{fig:farm}
\end{figure}

\begin{figure}[t]
\includegraphics[width=245pt,height=120pt]{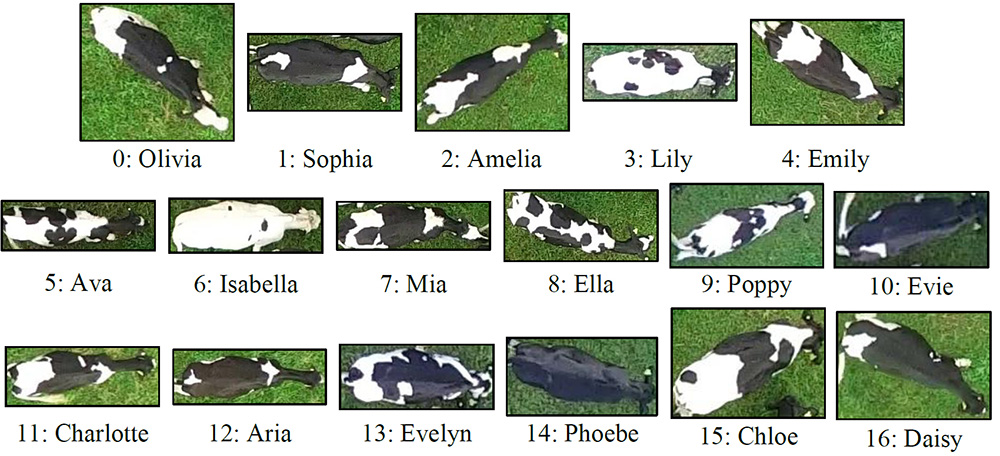}\vspace{-8pt}
\caption{\textbf{Target Cattle Herd.} 17 yearling heifer Holstein Friesian individuals comprising the full herd population used for experimentation throughout the paper. Note the uniqueness of their coat patterns utilised for biometric remote and tag-less identification by the robotic platform put forward.}\vspace{-10pt}
\label{fig:ind}
\end{figure}

Flights were performed over a two-week experimental period at the University of Bristol's Wyndhurst Farm in Langford Village, UK (see Fig.~\ref{fig:farm}) on a consistent herd of 17 yearling heifer Holstein Friesian cattle (see Fig.~\ref{fig:ind}), satisfying all relevant animal welfare and  flight regulations. 
Experiments consisted of two phases: \textit{(a)} training data acquisition across 14 (semi)manual flights, and subsequent \textit{(b)}~conduction of 18~autonomous flights.

\subsection{Training Data, Annotation and Augmentation}

Training data was acquired over two day-long recording sessions where manual and semi-autonomous flights at varying altitudes were carried out, recording video at a resolution of $3840 \times 2160$ at $30$fps.
The result was a raw dataset consisting of 37 minutes from 15~videos over 14 flights occupying 18GB. 
Overall $2285$~frames were extracted from these video files at a rate of $1 Hz$. 

\begin{figure}[b]
\includegraphics[width=245pt,height=205pt]{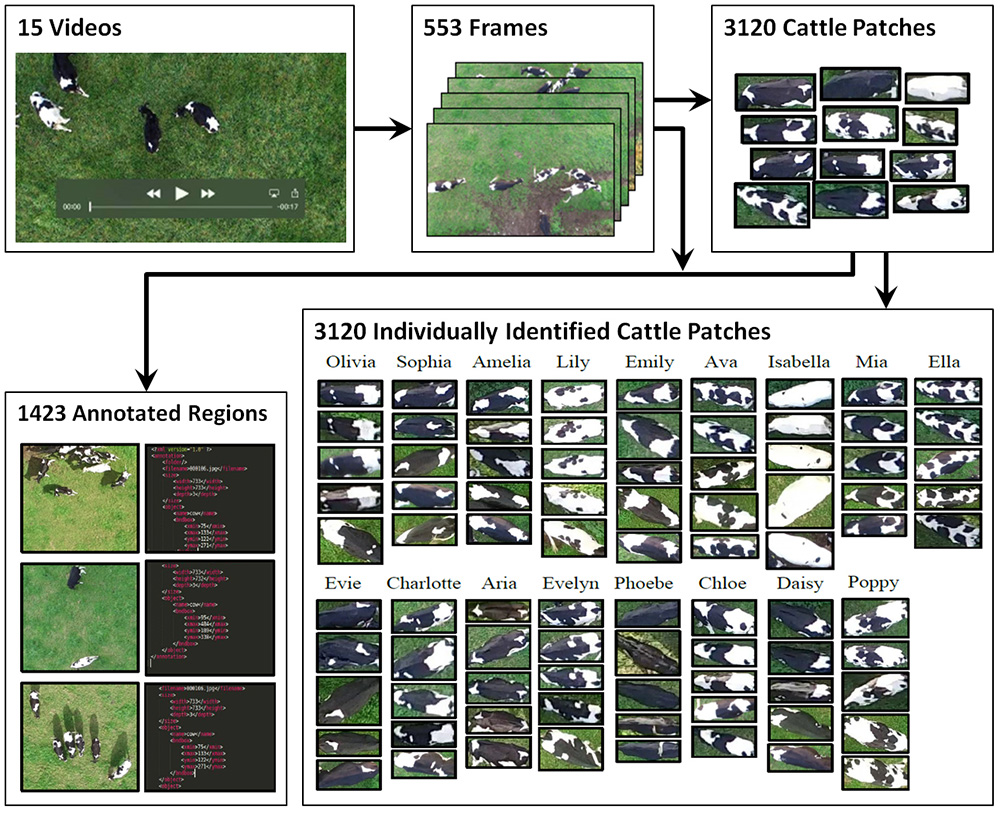}\vspace{-8pt}
\caption{\textbf{Training Data Annotation.} Across all 15 videos gathered for training data acquisition, $553$ frames contained cattle. Labelling bounding boxes around individual animals yielded $3120$ cattle patches. Labelling $720\times 720$ regions containing annotated cattle patches yielded the $1423$ images annotated with cattle bounding boxes that form the base data for  species detector training. Manual identification of cows, on the other hand, produced the base data for training the individual identification components.}
\label{fig:herd}
\end{figure}

\begin{figure}[t]
\vspace{4pt}
\includegraphics[width=0.485\textwidth]{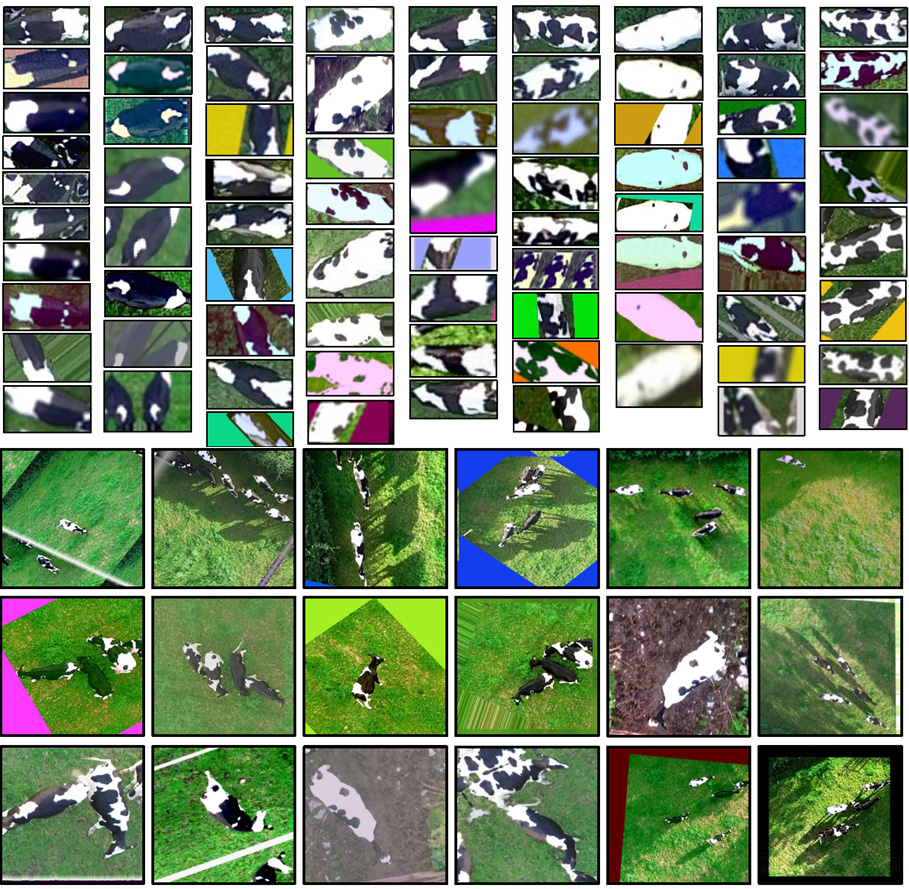}\vspace{-8pt}
\caption{\textbf{Data Augmentation.} \textit{\textbf{(top)}} For each of the first 9 identities out of 17, the top image per column shows a random non-synthetic example whilst other images below depict sample augmentations as used in stacks of 5 images for the  training of the LRCN based on Inception V3~\cite{szegedy2016rethinking}. \textbf{\textit{(bottom)}} Sample augmentations for  regions that -- together with augmented cattle bounding box annotations -- are provided as training input for the YOLOv2~\cite{redmon2017yolo9000} detection model.}\vspace{-15pt}
\label{fig:aug}
\end{figure}
First, after discarding frames without cattle, $3120$ bounding boxes around individual cattle were labelled in the $553$~frames containing cattle.
Animals were also manually identified as ground truth for individual identification. 
Secondly, to produce ground truth for training the cattle detector, square sub-images (matching the YOLOv2 input tensor size) were manually annotated to encompass individuals such that they are resolved at approximately $150\times 150$ pixels.
Figure~\ref{fig:herd} illustrates the full pipeline. 
To synthesise additional data, augmentations for both detection and identification datasets are performed stochastically with the possibility for any combination of the operations listed as follows according to a per-operation likelihood value: horizontal \& vertical flipping, crop \& pad, affine transformations (scale, translate, rotate, shear), Gaussian, average or median blurring, noise addition, background variations, contrast changes, and small perspective transformation. 
Figure~\ref{fig:aug} provides augmentation~samples.


\section{Software, Implementation and Training}

\subsection{Object Class Detection}

Cattle detection and localisation is performed  in real-time frame-by-frame using the YOLOv2~\cite{redmon2017yolo9000} CNN.
The network was retrained from scratch on the annotated region dataset, consisting of $11,384$~synthetic and non-synthetic training images (see Fig.~\ref{fig:aug}, bottom) and associated ground truth labels. 
Model inference operates on $736\times 736$ images obtained by cropping and scaling the source $960 \times 720$ pixel camera stream. 
As shown in Figure~\ref{fig:overview}, this process yields a set of $m$ bounding boxes $B=\{b_0, b_1, ..., b_{m-1}\}$ per frame with associated object confidence scores.
Inference on each of $n=5$ sampled frames  then produces  a box-annotated spatio-temporal  volume~$\{B_0, B_1, ..., B_{n-1}\}$.
Bounding boxes are associated across this volume by accumulating detections that are consistently present in cow-sized areas of an equally subdivided image.
This method is effective for reliable short-term tracking due to distinct and non-overlapping targets, slow target movement and stable UAV hovering. 
The outputs are $p\leq n$ short individual animal tracklets reshaped into an image patch sequence~$\{T_0, T_1, ..., T_p\}$, which forms the input to the individual identification network (see Section \ref{subsec:identity-estimation}). 
In addition, the current frame $I$ is also abstracted to a $5\times 5$ grid map $S$ encoding animal presence in the field of view of the camera (see Fig.~\ref{fig:overview}).
This forms the input to the exploratory agency network discussed in the following section.

\subsection{Exploratory Agency $(EA)$}
\label{subsec:exploratory-agency}
\vspace{-1pt}

Navigation activities aim at locating as many -- themselves moving -- individual animals as possible on the shortest routes in a gridded domain where a target counts as `located' once the agent occupies the same grid location as the target. 
To solve this dynamic travelling salesman task with  city locations to be discovered on the fly, we use a dual-stream deep network architecture, as first suggested in our previous work \cite{andrew2018deep}.
The method computes grid-based navigational decisions~$\{N,W,S,E\}$ based on immediate sensory~(tactical/exploitation) and historic navigational (strategic/exploration) information using two separate streams within a \textit{single} deep inference network. 
As shown in the paper, this strategy can significantly outperform simple strategies such as a `lawnmower' pattern and other baselines. 

To summarise the method's operation, the sensory input~$S$ is processed via a first stream utilising a basic AlexNet~\cite{krizhevsky2012imagenet} design (see Fig.~\ref{fig:overview}). 
A second stream operates on the exploratory history thus far, as stored in a long-term memory map $M$ (see Fig.~\ref{fig:overview}).
This stores the agent's present and past positions alongside animal encounters within the flight area of $20\times 20$~grid locations. 
The agent's starting position is fixed and $M$ is reset after $\delta\%$ of the map has been explored. 
Both these streams are concatenated into a shallow integration network that, as shown in Figure~\ref{fig:overview}, maps to a SoftMax-normalised likelihood vector~$V$ of the possible navigational actions. 
During inference, the network selects the top-ranking navigational action from~$\{N,W,S,E\}$ based
on~$V$, which is performed and, in-turn, the positional history~$M$ is updated. 

\begin{figure}[b]
\includegraphics[width=245pt,height=159pt]{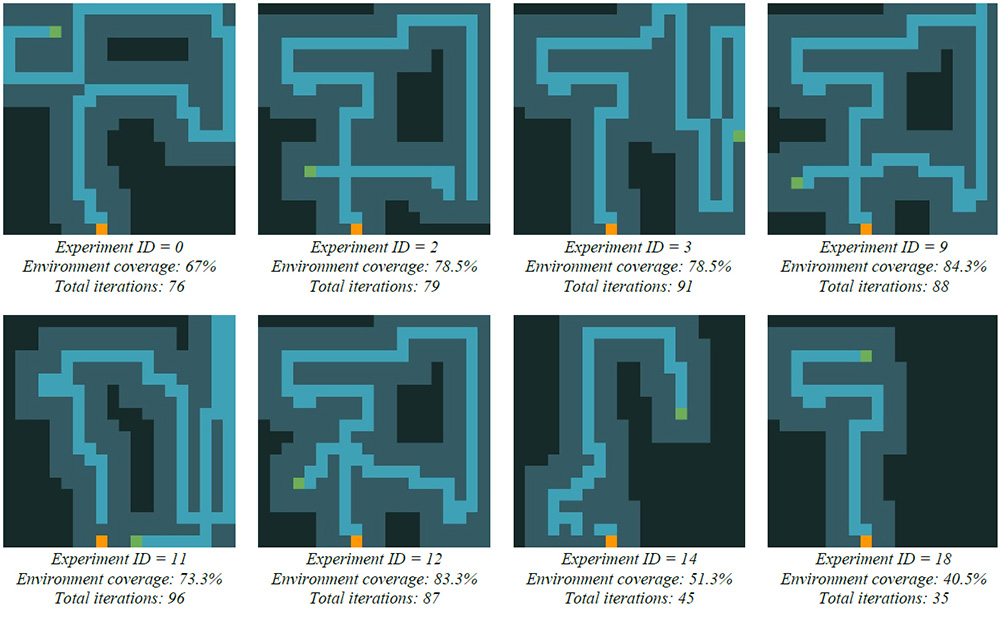}\vspace{-8pt}
\caption{\textbf{Autonomous Environment Explorations.} Examples of  8 autonomous aircraft paths of varying length (due to different battery charge statuses) chosen by the dual-stream navigational network trying to detect new moving targets quickly. Depictions show the $20\times20$ exploratory grid with agent local sensing of $5\times5$ cells for a $720\times720$ pixel image operating at a height of $10$ meters. The experiment starting (orange) and finishing (green) grid cell are also labelled. Note that the determinism of the exploratory
agency architecture is visible within instances where no target was detected; the bottom-right most exploration
(experiment ID = 18) depicts this baseline movement pattern.}
\label{fig:paths}
\end{figure}

\begin{figure*}[t]
\includegraphics[width=183pt,height=170pt]{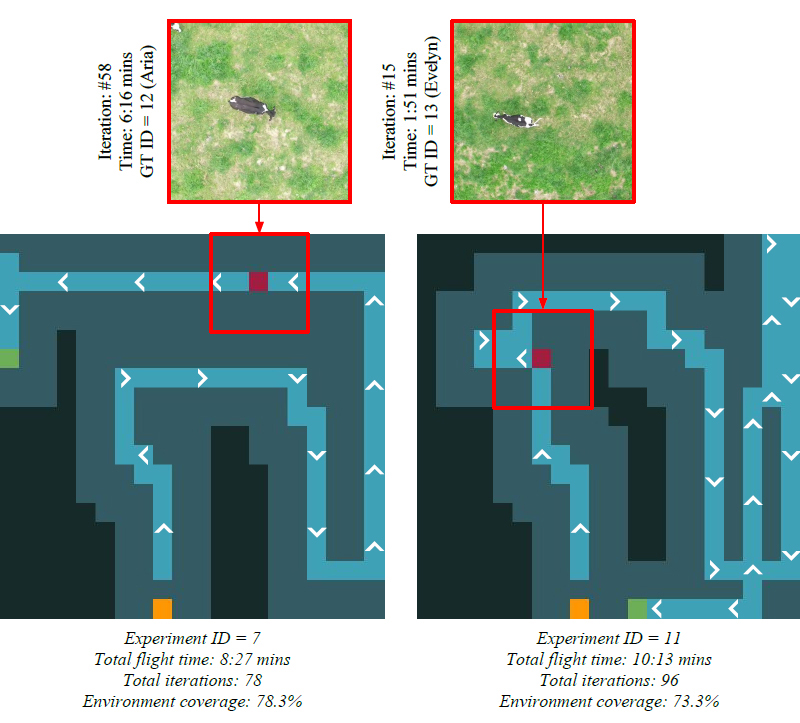}\ \ \includegraphics[width=315pt,height=164pt]{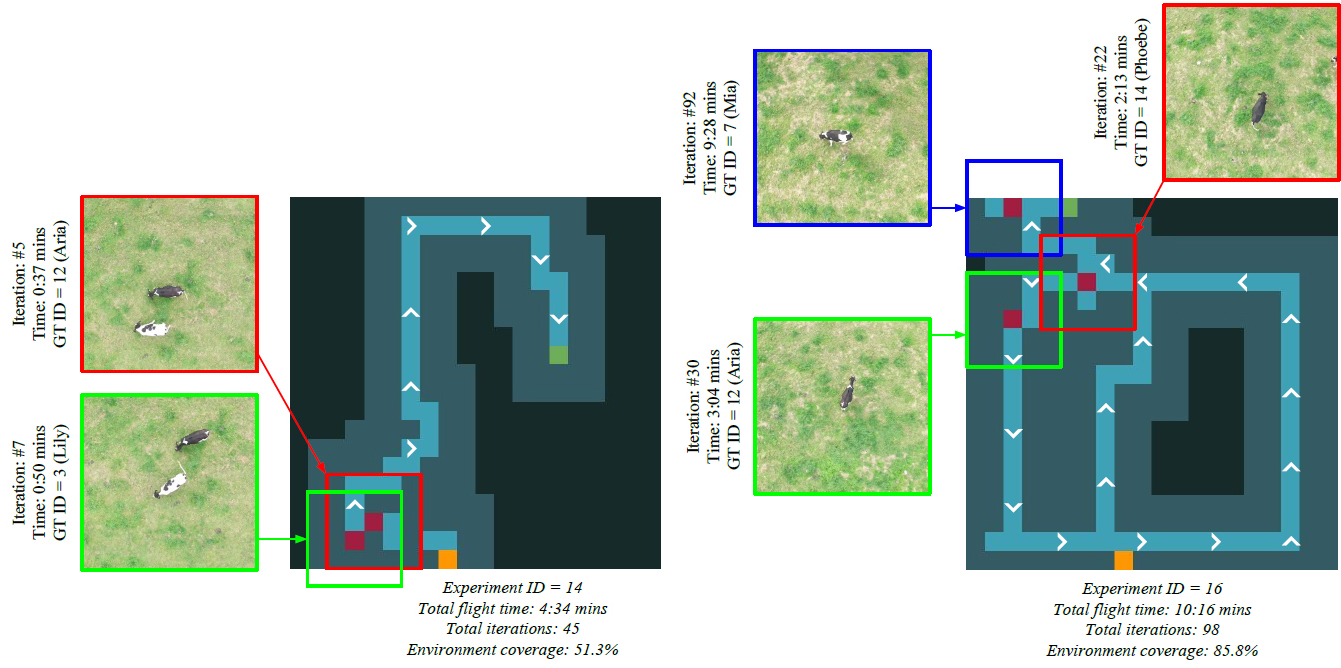} \vspace{-18pt}
\caption{\textbf{Annotated Autonomous Flights.} Examples of annotated agent flight paths within the exploratory
grid over the entire course of the experiment. 
Cell colours are defined as (black): unvisited locations, (light blue): visited locations, (dark blue): seen or covered areas, (orange): agent starting position, (green): finishing agent position and (red): discovered target positions. 
At each target discovery point the corresponding captured image $I$ is shown with statistics. 
Also illustrated by white arrows is the agent's direction of travel at iteration intervals throughout the experiment. 
Across every experiment, the average time required per iteration is: $6.35$ seconds including movement to the location from a neighbouring cell and performing identification if required.}\vspace{-10pt}
\label{fig:flights}
\end{figure*}

For training, the entire two-stream navigation network is optimised via stochastic gradient decent (SGD) with momentum \cite{qian1999momentum} and a fixed learning rate $e = 0.001$ based on triples~$(S;M;V)$ using one-hot encoding of~$V$ and cross-entropy loss. 
This unified model allows for the back-propagation of navigation decision errors across \textit{both} streams and domains. 
For training, we simulate $10,000$ episodes of $17$ pseudo-randomly \cite{matsumoto1998mersenne} placed targets in a $20\times 20$ grid and calculate optimal navigation decisions $(S;M;V)$ by solving the associated travelling salesman problem.
$10$-fold cross validation on this setup yielded an accuracy of $72.45\%$ in making an optimal next grid navigation decision and a target recovery rate of $0.26 \pm 0.06$ targets per grid~move. 
For full implementation details we refer to the original paper~\cite{andrew2018deep}, which operates on simulations. In contrast, examples of real-world environment explorations during our 18 test flights are visualised in Figure~\ref{fig:paths} and Figure~\ref{fig:flights}. 

\subsection{Coordinate Fulfilment}

Re-positioning commands from~$\{N,W,S,E\}$ to the M100 flight platform need to be issued via local position offsets in metres with respect to a programatically-set East North Up~(ENU) reference frame.
As such, in order to fulfil a target GPS coordinate arising from exploratory agency, it must be converted into that  frame.
This is achieved by converting the target GPS coordinate into the static Earth-Centred Earth-Fixed (ECEF) reference frame, then converting that coordinate into the local ENU frame.
Equally, the same process is performed on the agent's current GPS position and the resulting local positions are compared.
Our implementation follows the  standard as established in the literature \cite{barth1999global, hofmann2001gps}.

\subsection{Identity Estimation $(IE)$}
\label{subsec:identity-estimation}

Individual identification based on an image patch sequence~$\{T_0, T_1, ..., T_p\}$ is performed via an LRCN, first introduced by Donahue et al. \cite{donahue2015long}.
In particular, as shown in Figure~\ref{fig:overview}, we combine a GoogLeNet/Inception V3 CNN~\cite{szegedy2015going,szegedy2016rethinking}  with a single Long Short-Term Memory (LSTM) \cite{hochreiter1997long} layer. 
This approach has demonstrated success in disambiguating fine-grained categories in our previous work~\cite{andrew2017visual}.

\begin{figure}[hb]
\includegraphics[width=174pt,height=145pt]{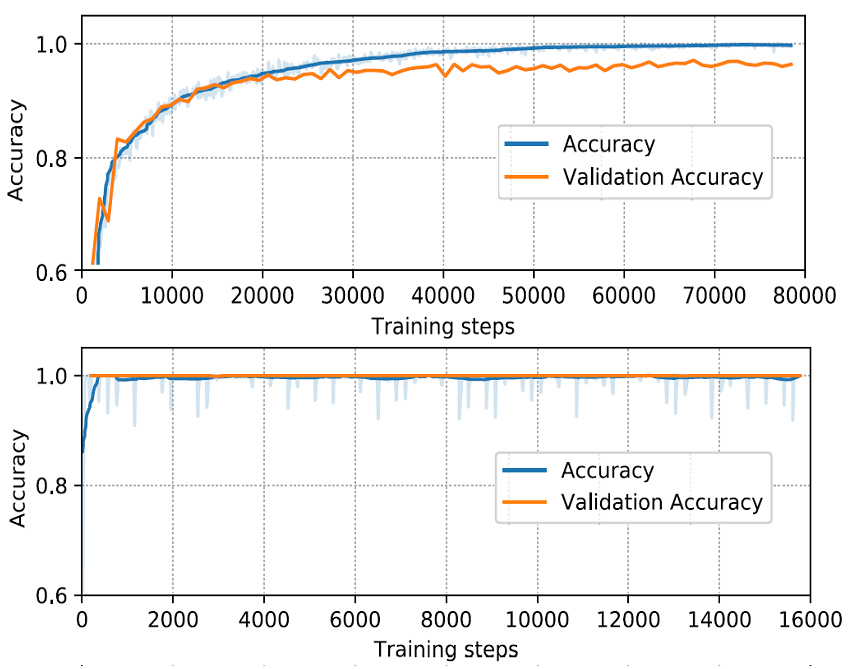}\includegraphics[width=72pt,height=143pt]{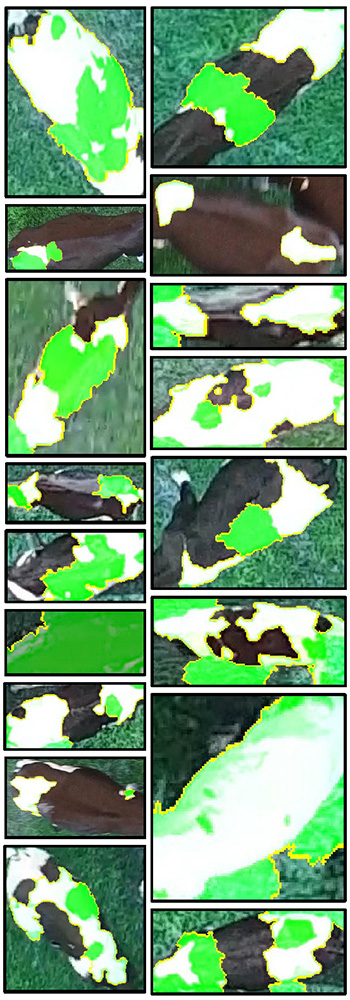}\vspace{-8pt}
\caption{\textbf{Training Individual Identification Components.} \textit{\textbf{(top)}} Training and validation accuracies versus training steps for the Inception V3 architecture~\cite{szegedy2016rethinking} smoothed for visualisation purposes using the Savitzky-Golay filter~\cite{savitzky1964smoothing} with value capping at 1. 
This approach yields 97.13\% identification accuracy when operating on a single $224\times 224$ input image patch of a cow.
\textbf{\textit{(Bottom)}} The proposed LRCN architecture operates on $5$ class-identical such patches yielding perfect validation performance.
\textbf{\textit{(Right)}} Hand-picked local interpretable model-agnostic explanations~\cite{ribeiro2016should} from the single image Inception/GoogLeNet approach for each possible cow identity illustrating the learning of per-category discriminative features. 
Green regions depict the superpixel(s) that were activated for correct predictions.}\vspace{-7pt}
\label{fig:train}
\end{figure}

Training of the GoogLeNet/Inception V3 network takes groups of  $n=5$ same class randomly selected RoIs (exemplified in Fig.~\ref{fig:aug}, top), each of which were non-proportionally resized to $224 \times 224$ pixels. 
SGD with momentum \cite{qian1999momentum}, a batch size of 32 and a fixed learning rate $e = 0.001$ were used for optimisation. 
Figure \ref{fig:train} \textit{(right)} provides evidence of per-category learning of appropriate spatial representations using local interpretable model-agnostic explanations~\cite{ribeiro2016should}, which qualitatively highlight the success of the Inception architecture learning discriminative and fine-grained visual features for each individual. 
 Once trained, samples are passed through this GoogLeNet up to the pool\_5 layer and feature vectors are combined over the~$n$ samples.
A shallow LSTM network is finally trained on these vector sequences using a SoftMax cross-entropy cost function optimised  against the one-hot encoded identities vector representing the $17$ possible classes. 
This approach achieved 100\% validation accuracy with little training, as can be seen in Figure \ref{fig:train}~bottom.\vspace{-5pt}

\section{Real-World Autonomous Performance}

We conducted  $18$ fully autonomous test flights at a low altitude (approximately $10$m) above an area of $20\times 20$ cells~(see Fig.~\ref{fig:farm}) covering altogether 147~minutes. 
Examples of environment explorations are visualised in Figure~\ref{fig:paths} and Figure~\ref{fig:flights} depicting various example flights with detailed annotations of flight path, animal encounters and identification confidences.  
For all experiments, we ran the object detection and exploratory agency networks live and in real-time to navigate the aircraft. 
Note that the herd was naturally dispersed in these experiments and animals were free to roam across the entire field in and out of the covered area. 
Thus, only a few individuals were present and recoverable in this area at any one time. 
The median coverage of the  grid was  $70.13\%$ with median flight time of $8$ minutes and $9$~seconds per experiment, and a median of $77$ grid iterations per flight. 
For each of the flights, we conducted two types of experiment: that is \textit{(a)} saving a single frame per grid location (due to onboard storage limitations) visited and perform a full separate analysis of detection and identification performance after the flight \textit{offline}, and \textit{(b)} also running multi-frame identification live during flight for all cases where the aircraft has navigated centrally above a detected individual.

\subsection{Offline After-Flight Performance Evaluation}

\begin{figure}[t]
 \vspace{6pt}
\includegraphics[width=245pt,height=125pt]{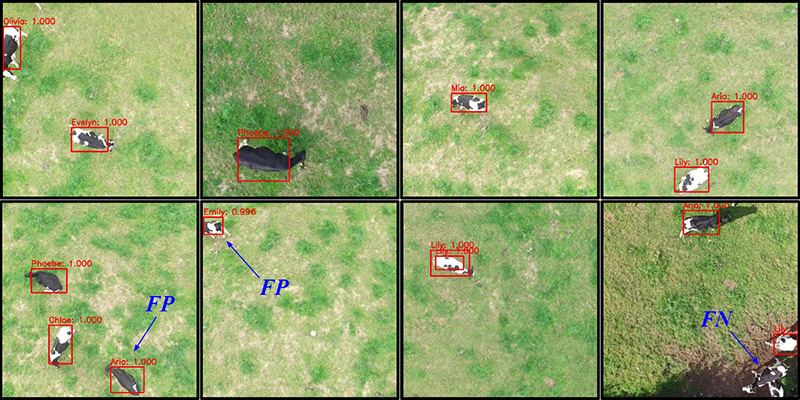}\vspace{-8pt}
\caption{\textbf{Examples of Identification Success and Failure.} \textit{\textbf{(top row)}}~Examples of detection and identification successes where red boxes denote  YOLOv2 detections with predicted identity and confidence values. \textit{\textbf{(bottom row)}}~Failures  from left to right: false positive due to pattern similarity; false positive due to instance cropping; nested double detection; false negative.}\vspace{3pt}
\label{fig:offline}
\end{figure}

For \textit{offline} analysis, the UAV saved to file one acquired $720 \times 720$ image at each exploratory agency iteration, yielding $1039$ images.
$99$ of those images actually contained target cows that were hand labelled with ground truth bounding box annotations and identities according to the VOC guidelines \cite{everingham2010pascal}.
A detection was deemed a successful true positive based on the IoU ($ov \geq 0.5$). 
Grounded in this, we measured the YOLOv2 detection accuracy to be $92.4\%$, where out of the $111$ present animals, $2$ were missed and $7$ false positive nested detections occurred (see Fig.~\ref{fig:offline}).
We then tested separately, the performance of the single frame Inception V3 individual identification architecture (yielding $93.6\%$ accuracy), where all ground truth bounding boxes (not only the detected instances) were presented to the ID component. 
In contrast, and as shown in Table~\ref{table:offline}, when identification is performed on detected RoIs only then the combined offline system accuracy is 91.9\%.

\subsection{Online In-Flight Performance Evaluation}
\begin{figure}[b]
\includegraphics[width=75pt,height=110pt]{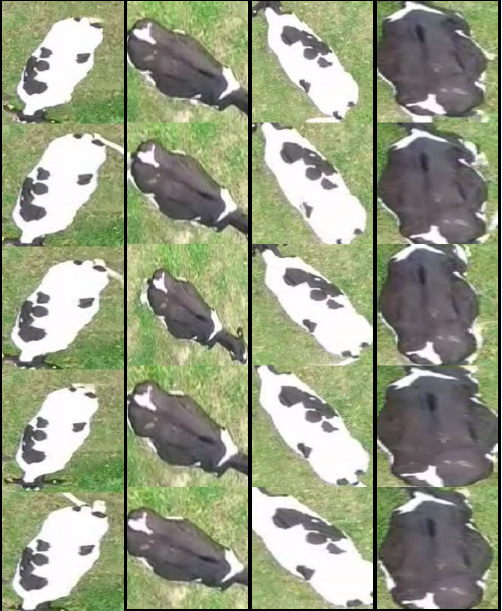}\ \includegraphics[width=165pt,height=110pt]{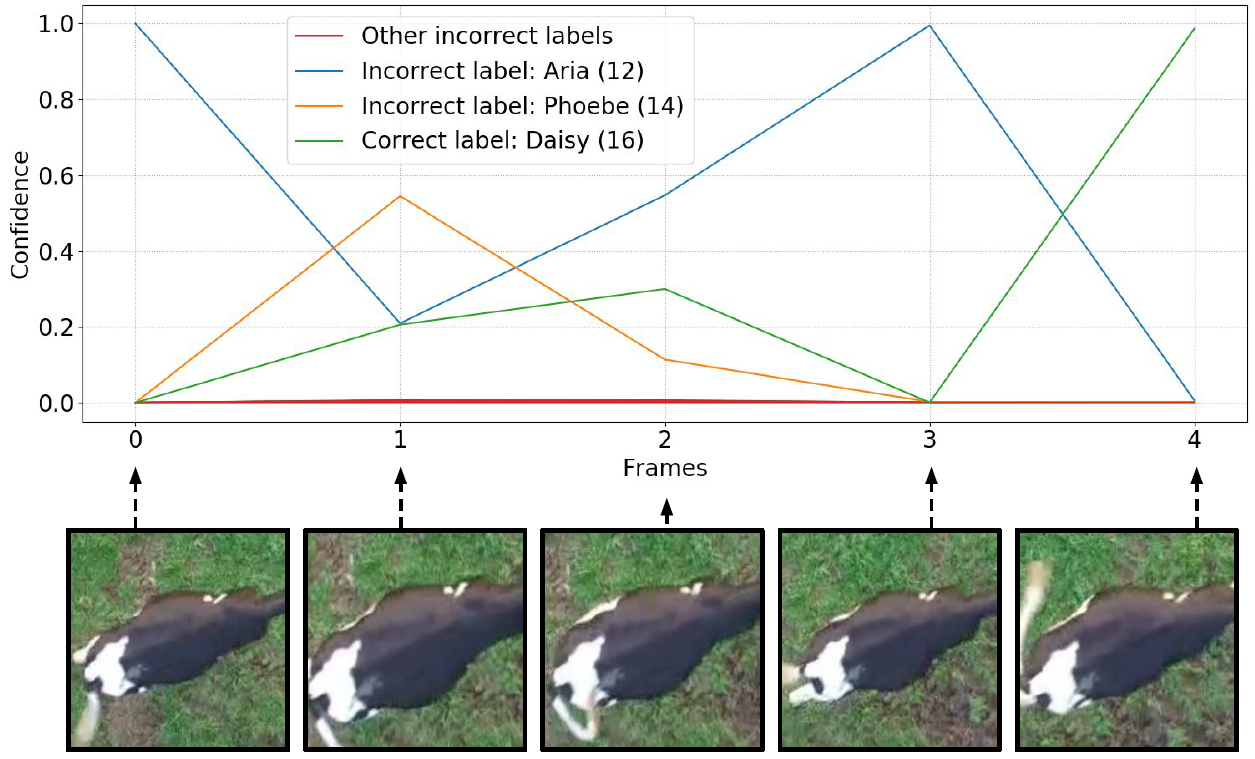}\vspace{-8pt}
\caption{\textbf{Effectiveness of Temporal  Integration of Evidence.} \textit{\textbf{(left)}}~Examples of $224\times224$ pixel same-class RoI sequences~$T$ (tracklets) at temporal length $5$ as provided to the LRCN component. \textit{\textbf{(right)}} An example where LRCN model confidence for the single
correct and other incorrect identities are plotted versus exposure to subsequent frames. Initial false identification is overcome by exposure to new views. Note that movement of the animal's tail impacts on feature appearance and leads initially to erroneous
visual similarities comparable to Aria (ID=12). }\vspace{-2pt}
\label{fig:temporal}
\end{figure}

For \textit{online} autonomous operation, all computation was performed live in real-time onboard the UAV's computers~(DJI Manifold \& Nvidia Jetson TX2).
Figure~\ref{fig:flights} depicts various example flights with detailed annotations of flight paths, animal encounters and identification confidences. Across the $1039$ grid locations visited during the set of experiments, the aircraft  navigated centrally above a detected individual $18$ times triggering identification. 
Note that this mode of operation eliminates the problem of clipped visibility at image borders, minimises image distortions, optimises the viewpoint, and exposes the coat in a canonized orthogonal view. 
For triggered identification, we store intermediate LRCN confidence outputs after processing up to $5$ same-class patches $T$ to compare performance differences between single view and multi-view identification.  
Figure~\ref{fig:temporal} depicts some same-class patch sequences~$T$ and one instance where multi-frame inference was indeed beneficial to identification.
The respective overall results are given in Table \ref{table:7-full:online-results}.
Notably, across the small online sample set ($18$  instances), the LRCN model performs~perfectly.

\begin{table}[t]
\centering \begin{scriptsize}
\resizebox{\columnwidth}{!}{%
\begin{tabular}{|c|c|c|c|c|}
\hline
&& Detection on &  Single Frame ID &Combined\\
\# Sample&\# Animal& Sample Frames &  on Labelled RoIs &Detection+ID\\
Frames&Instances&Accuracy (\%) & Accuracy  (\%) & Accuracy  (\%) \\ \hline \hline
1039&111&92.4          & 93.6  & 91.9         \\
\hline
\end{tabular}}\end{scriptsize}
\caption{\textbf{Offline Performance Results.}}\vspace{-15pt}
\label{table:offline}
\end{table}
\begin{table}[hb]
\centering
\resizebox{\columnwidth}{!}{%
\begin{tabular}{|c||c|c|}
\hline
           & LRCN Identification & Single Frame Identification \\ 
\# Samples &  Accuracy (\%)                       &  Accuracy (\%)                   \\ \hline \hline
18                       & \textbf{100}                 &      94.4                           \\ \hline
\end{tabular}}
\caption{\textbf{Online Identification Results.}}\vspace{-25pt}
\label{table:7-full:online-results}
\end{table}

\vspace{11pt}
\section{Conclusion and Future Work}
This paper provides a proof-of-concept that fully autonomous aerial animal biometrics is practically feasible. Operating in a real-world agricultural setting, the paper demonstrated that individual cattle identities can be reliably  recovered biometrically from the air onboard a fully autonomous robotic agent{. 
Experiments conducted on a small herd of 17 live cattle confirmed demonstrable identification robustness of the proposed approach. 
In successfully performing these tasks with limited computational resources alongside payload, weight restrictions and more, the presented system gives rise to future agricultural automation possibilities with potential positive implications for animal welfare and farm productivity.

Beyond farming, the concept of  autonomous biometric animal identification from the air  as presented  opens up a realm of future applications in  fields such as ecology, where animal identification of uniquely patterned species in the wild~(e.g. zebras, giraffes) is critical to assessing the status of  populations.

\balance
\bibliographystyle{IEEEtran}
\bibliography{references}

\begin{thebibliography}{10}
\providecommand{\url}[1]{#1}
\csname url@rmstyle\endcsname
\providecommand{\newblock}{\relax}
\providecommand{\bibinfo}[2]{#2}
\providecommand\BIBentrySTDinterwordspacing{\spaceskip=0pt\relax}
\providecommand\BIBentryALTinterwordstretchfactor{4}
\providecommand\BIBentryALTinterwordspacing{\spaceskip=\fontdimen2\font plus
\BIBentryALTinterwordstretchfactor\fontdimen3\font minus
  \fontdimen4\font\relax}
\providecommand\BIBforeignlanguage[2]{{%
\expandafter\ifx\csname l@#1\endcsname\relax
\typeout{** WARNING: IEEEtran.bst: No hyphenation pattern has been}%
\typeout{** loaded for the language `#1'. Using the pattern for}%
\typeout{** the default language instead.}%
\else
\language=\csname l@#1\endcsname
\fi
#2}}

\bibitem{ungerfeld2014time}
R.~Ungerfeld, C.~Cajarville, M.~Rosas, and J.~Repetto, ``Time budget
  differences of high-and low-social rank grazing dairy cows,'' \emph{New
  Zealand journal of agricultural research}, vol.~57, no.~2, pp. 122--127,
  2014.

\bibitem{kondo1990stabilization}
S.~Kondo and J.~Hurnik, ``Stabilization of social hierarchy in dairy cows,''
  \emph{Applied Animal Behaviour Science}, vol.~27, no.~4, pp. 287--297, 1990.

\bibitem{phillips2002effects}
C.~Phillips and M.~Rind, ``The effects of social dominance on the production
  and behavior of grazing dairy cows offered forage supplements,''
  \emph{Journal of Dairy Science}, vol.~85, no.~1, pp. 51--59, 2002.

\bibitem{gregorini2012diurnal}
P.~Gregorini, ``Diurnal grazing pattern: its physiological basis and strategic
  management,'' \emph{Animal Production Science}, vol.~52, no.~7, pp. 416--430,
  2012.

\bibitem{gregorini2006behavior}
P.~Gregorini, S.~Tamminga, and S.~Gunter, ``Behavior and daily grazing patterns
  of cattle,'' \emph{The Professional Animal Scientist}, vol.~22, no.~3, pp.
  201--209, 2006.

\bibitem{sowell2000social}
B.~Sowell, J.~Mosley, and J.~Bowman, ``Social behavior of grazing beef cattle:
  Implications for management,'' \emph{Journal of Animal Science}, vol.~77, no.
  E-Suppl, pp. 1--6, 2000.

\bibitem{lin2009uav}
L.~Lin and M.~A. Goodrich, ``Uav intelligent path planning for wilderness
  search and rescue,'' in \emph{Intelligent robots and systems, 2009. IROS
  2009. IEEE/RSJ International Conference on}.\hskip 1em plus 0.5em minus
  0.4em\relax IEEE, 2009, pp. 709--714.

\bibitem{waharte2010supporting}
S.~Waharte and N.~Trigoni, ``Supporting search and rescue operations with
  uavs,'' in \emph{Emerging Security Technologies (EST), 2010 International
  Conference on}.\hskip 1em plus 0.5em minus 0.4em\relax IEEE, 2010, pp.
  142--147.

\bibitem{ryu2012autonomous}
K.~Ryu, ``Autonomous robotic strategies for urban search and rescue,'' Ph.D.
  dissertation, Virginia Polytechnic Institute and State University, 2012.

\bibitem{bonin2008visual}
F.~Bonin-Font, A.~Ortiz, and G.~Oliver, ``Visual navigation for mobile robots:
  A survey,'' \emph{Journal of intelligent and robotic systems}, vol.~53,
  no.~3, p. 263, 2008.

\bibitem{desouza2002vision}
G.~N. DeSouza and A.~C. Kak, ``Vision for mobile robot navigation: A survey,''
  \emph{IEEE transactions on pattern analysis and machine intelligence},
  vol.~24, no.~2, pp. 237--267, 2002.

\bibitem{matsumoto1996visual}
Y.~Matsumoto, M.~Inaba, and H.~Inoue, ``Visual navigation using view-sequenced
  route representation,'' in \emph{Robotics and Automation, 1996. Proceedings.,
  1996 IEEE International Conference on}, vol.~1.\hskip 1em plus 0.5em minus
  0.4em\relax IEEE, 1996, pp. 83--88.

\bibitem{jones1997appearance}
S.~D. Jones, C.~Andresen, and J.~L. Crowley, ``Appearance based process for
  visual navigation,'' in \emph{Intelligent Robots and Systems, 1997. IROS'97.,
  Proceedings of the 1997 IEEE/RSJ International Conference on}, vol.~2.\hskip
  1em plus 0.5em minus 0.4em\relax IEEE, 1997, pp. 551--557.

\bibitem{Santos1993divergent}
J.~Santos-Victor, G.~Sandini, F.~Curotto, and S.~Garibaldi, ``Divergent stereo
  for robot navigation: Learning from bees,'' in \emph{Computer Vision and
  Pattern Recognition, 1993. Proceedings CVPR'93., 1993 IEEE Computer Society
  Conference on}.\hskip 1em plus 0.5em minus 0.4em\relax IEEE, 1993, pp.
  434--439.

\bibitem{pears2001ground}
N.~Pears and B.~Liang, ``Ground plane segmentation for mobile robot visual
  navigation,'' in \emph{Intelligent Robots and Systems, 2001. Proceedings.
  2001 IEEE/RSJ International Conference on}, vol.~3.\hskip 1em plus 0.5em
  minus 0.4em\relax IEEE, 2001, pp. 1513--1518.

\bibitem{saeedi2006vision}
P.~Saeedi, P.~D. Lawrence, and D.~G. Lowe, ``Vision-based 3-d trajectory
  tracking for unknown environments,'' \emph{IEEE transactions on robotics},
  vol.~22, no.~1, pp. 119--136, 2006.

\bibitem{giusti2016machine}
A.~Giusti, J.~Guzzi, D.~C. Cire{\c{s}}an, F.-L. He, J.~P. Rodr{\'\i}guez,
  F.~Fontana, M.~Faessler, C.~Forster, J.~Schmidhuber, G.~Di~Caro,
  \emph{et~al.}, ``A machine learning approach to visual perception of forest
  trails for mobile robots,'' \emph{IEEE Robotics and Automation Letters},
  vol.~1, no.~2, pp. 661--667, 2016.

\bibitem{ran2017convolutional}
L.~Ran, Y.~Zhang, Q.~Zhang, and T.~Yang, ``Convolutional neural network-based
  robot navigation using uncalibrated spherical images,'' \emph{Sensors},
  vol.~17, no.~6, p. 1341, 2017.

\bibitem{andrew2018deep}
W.~Andrew, C.~Greatwood, and T.~Burghardt, ``Deep learning for exploration and
  recovery of uncharted and dynamic targets from uav-like vision,'' in
  \emph{2018 IEEE/RSJ International Conference on Intelligent Robots and
  Systems (IROS)}.\hskip 1em plus 0.5em minus 0.4em\relax IEEE, 2018, pp.
  1124--1131.

\bibitem{borenstein1990real}
J.~Borenstein and Y.~Koren, ``Real-time obstacle avoidance for fast mobile
  robots in cluttered environments,'' in \emph{Robotics and Automation, 1990.
  Proceedings., 1990 IEEE International Conference on}.\hskip 1em plus 0.5em
  minus 0.4em\relax IEEE, 1990, pp. 572--577.

\bibitem{oriolo1995line}
G.~Oriolo, M.~Vendittelli, and G.~Ulivi, ``On-line map building and navigation
  for autonomous mobile robots,'' in \emph{Robotics and Automation, 1995.
  Proceedings., 1995 IEEE International Conference on}, vol.~3.\hskip 1em plus
  0.5em minus 0.4em\relax IEEE, 1995, pp. 2900--2906.

\bibitem{moravec1983stanford}
H.~P. Moravec, ``The stanford cart and the cmu rover,'' \emph{Proceedings of
  the IEEE}, vol.~71, no.~7, pp. 872--884, 1983.

\bibitem{kosaka1992fast}
A.~Kosaka and A.~C. Kak, ``Fast vision-guided mobile robot navigation using
  model-based reasoning and prediction of uncertainties,'' \emph{CVGIP: Image
  understanding}, vol.~56, no.~3, pp. 271--329, 1992.

\bibitem{Zhang2017neural}
J.~Zhang, L.~Tai, J.~Boedecker, W.~Burgard, and M.~Liu, ``Neural slam,''
  \emph{arXiv preprint arXiv:1706.09520}, 2017.

\bibitem{melnikov2014projective}
A.~A. Melnikov, A.~Makmal, and H.~J. Briegel, ``Projective simulation applied
  to the grid-world and the mountain-car problem,'' \emph{arXiv preprint
  arXiv:1405.5459}, 2014.

\bibitem{kuhl2013animal}
H.~S. K{\"u}hl and T.~Burghardt, ``Animal biometrics: quantifying and detecting
  phenotypic appearance,'' \emph{Trends in ecology \& evolution}, vol.~28,
  no.~7, pp. 432--441, 2013.

\bibitem{lowe1999object}
D.~G. Lowe, ``Object recognition from local scale-invariant features,'' in
  \emph{Computer vision, 1999. The proceedings of the seventh IEEE
  international conference on}, vol.~2.\hskip 1em plus 0.5em minus 0.4em\relax
  Ieee, 1999, pp. 1150--1157.

\bibitem{martinez2013video}
C.~A. Martinez-Ortiz, R.~M. Everson, and T.~Mottram, ``Video tracking of dairy
  cows for assessing mobility scores,'' 2013.

\bibitem{morel2009asift}
J.-M. Morel and G.~Yu, ``Asift: A new framework for fully affine invariant
  image comparison,'' \emph{SIAM Journal on Imaging Sciences}, vol.~2, no.~2,
  pp. 438--469, 2009.

\bibitem{Andrew2016automatic}
W.~Andrew, S.~Hannuna, N.~Campbell, and T.~Burghardt, ``Automatic individual
  holstein friesian cattle identification via selective local coat pattern
  matching in rgb-d imagery,'' in \emph{Image Processing (ICIP), 2016 IEEE
  International Conference on}.\hskip 1em plus 0.5em minus 0.4em\relax IEEE,
  2016, pp. 484--488.

\bibitem{andrew2017visual}
W.~Andrew, C.~Greatwood, and T.~Burghardt, ``Visual localisation and individual
  identification of holstein friesian cattle via deep learning,'' in \emph{2017
  IEEE International Conference on Computer Vision Workshops (ICCVW)}, Oct
  2017, pp. 2850--2859.

\bibitem{donahue2015long}
J.~Donahue, L.~Anne~Hendricks, S.~Guadarrama, M.~Rohrbach, S.~Venugopalan,
  K.~Saenko, and T.~Darrell, ``Long-term recurrent convolutional networks for
  visual recognition and description,'' in \emph{Proceedings of the IEEE
  conference on computer vision and pattern recognition}, 2015, pp. 2625--2634.

\bibitem{hodgson2013unmanned}
A.~Hodgson, N.~Kelly, and D.~Peel, ``Unmanned aerial vehicles (uavs) for
  surveying marine fauna: a dugong case study,'' \emph{PloS one}, vol.~8,
  no.~11, p. e79556, 2013.

\bibitem{koski2009evaluation}
W.~R. Koski, T.~Allen, D.~Ireland, G.~Buck, P.~R. Smith, A.~M. Macrander, M.~A.
  Halick, C.~Rushing, D.~J. Sliwa, and T.~L. McDonald, ``Evaluation of an
  unmanned airborne system for monitoring marine mammals,'' \emph{Aquatic
  Mammals}, vol.~35, no.~3, p. 347, 2009.

\bibitem{abd2005development}
A.~Abd-Elrahman, L.~Pearlstine, and F.~Percival, ``Development of pattern
  recognition algorithm for automatic bird detection from unmanned aerial
  vehicle imagery,'' \emph{Surveying and Land Information Science}, vol.~65,
  no.~1, p.~37, 2005.

\bibitem{rodriguez2012eye}
A.~Rodr{\'\i}guez, J.~J. Negro, M.~Mulero, C.~Rodr{\'\i}guez,
  J.~Hern{\'a}ndez-Pliego, and J.~Bustamante, ``The eye in the sky: combined
  use of unmanned aerial systems and gps data loggers for ecological research
  and conservation of small birds,'' \emph{PLoS One}, vol.~7, no.~12, p.
  e50336, 2012.

\bibitem{sloopflyer}
MIT, ``Sloopflyer,'' https://caos.mit.edu/blog/glider-photography-sloopflyer,
  [Online; accessed 1-Mar-2019. Unpublished elsewhere].

\bibitem{hui2018vision}
X.~Hui, J.~Bian, X.~Zhao, and M.~Tan, ``Vision-based autonomous navigation
  approach for unmanned aerial vehicle transmission-line inspection,''
  \emph{International Journal of Advanced Robotic Systems}, vol.~15, no.~1, p.
  1729881417752821, 2018.

\bibitem{cobo2016approach}
A.~F. Cobo and F.~C. Ben{\i}tez, ``Approach for autonomous landing on moving
  platforms based on computer vision,'' 2016.

\bibitem{yu2017intelligent}
H.~Yu, S.~Lin, J.~Wang, K.~Fu, and W.~Yang, ``{An Intelligent Unmanned Aircraft
  System for Wilderness Search and Rescue},''
  http://www.imavs.org/papers/2017/143{\_}imav2017{\_}proceedings.pdf, [Online;
  accessed 1-Mar-2019].

\bibitem{kyristsis2016towards}
S.~Kyristsis, A.~Antonopoulos, T.~Chanialakis, E.~Stefanakis, C.~Linardos,
  A.~Tripolitsiotis, and P.~Partsinevelos, ``Towards autonomous modular uav
  missions: The detection, geo-location and landing paradigm,'' \emph{Sensors},
  vol.~16, no.~11, p. 1844, 2016.

\bibitem{szegedy2016rethinking}
C.~Szegedy, V.~Vanhoucke, S.~Ioffe, J.~Shlens, and Z.~Wojna, ``Rethinking the
  inception architecture for computer vision,'' in \emph{Proceedings of the
  IEEE Conference on Computer Vision and Pattern Recognition}, 2016, pp.
  2818--2826.

\bibitem{redmon2017yolo9000}
\BIBentryALTinterwordspacing
J.~Redmon and A.~Farhadi, ``{YOLO9000:} better, faster, stronger,''
  \emph{CoRR}, vol. abs/1612.08242, 2016. [Online]. Available:
  \url{http://arxiv.org/abs/1612.08242}
\BIBentrySTDinterwordspacing

\bibitem{krizhevsky2012imagenet}
A.~Krizhevsky, I.~Sutskever, and G.~E. Hinton, ``Imagenet classification with
  deep convolutional neural networks,'' in \emph{Advances in neural information
  processing systems}, 2012, pp. 1097--1105.

\bibitem{qian1999momentum}
N.~Qian, ``On the momentum term in gradient descent learning algorithms,''
  \emph{Neural networks}, vol.~12, no.~1, pp. 145--151, 1999.

\bibitem{matsumoto1998mersenne}
M.~Matsumoto and T.~Nishimura, ``Mersenne twister: a 623-dimensionally
  equidistributed uniform pseudo-random number generator,'' \emph{ACM
  Transactions on Modeling and Computer Simulation (TOMACS)}, vol.~8, no.~1,
  pp. 3--30, 1998.

\bibitem{barth1999global}
M.~Barth and J.~A. Farrell, ``The global positioning system \& inertial
  navigation,'' \emph{McGraw-Hill}, vol.~8, pp. 21--56, 1999.

\bibitem{hofmann2001gps}
B.~Hofmann, H.~Lichtenegger, and J.~Collins, ``Gps theory and practice,''
  \emph{Springer Wien NewYork}, 2001.

\bibitem{szegedy2015going}
C.~Szegedy, W.~Liu, Y.~Jia, P.~Sermanet, S.~Reed, D.~Anguelov, D.~Erhan,
  V.~Vanhoucke, and A.~Rabinovich, ``Going deeper with convolutions,'' in
  \emph{Proceedings of the IEEE conference on computer vision and pattern
  recognition}, 2015, pp. 1--9.

\bibitem{hochreiter1997long}
S.~Hochreiter and J.~Schmidhuber, ``Long short-term memory,'' \emph{Neural
  computation}, vol.~9, no.~8, pp. 1735--1780, 1997.

\bibitem{ribeiro2016should}
M.~T. Ribeiro, S.~Singh, and C.~Guestrin, ``Why should i trust you?: Explaining
  the predictions of any classifier,'' in \emph{Proceedings of the 22nd ACM
  SIGKDD international conference on knowledge discovery and data
  mining}.\hskip 1em plus 0.5em minus 0.4em\relax ACM, 2016, pp. 1135--1144.

\bibitem{savitzky1964smoothing}
A.~Savitzky and M.~J. Golay, ``Smoothing and differentiation of data by
  simplified least squares procedures.'' \emph{Analytical chemistry}, vol.~36,
  no.~8, pp. 1627--1639, 1964.

\bibitem{everingham2010pascal}
M.~Everingham, L.~Van~Gool, C.~K. Williams, J.~Winn, and A.~Zisserman, ``The
  pascal visual object classes (voc) challenge,'' \emph{International journal
  of computer vision}, vol.~88, no.~2, pp. 303--338, 2010.

\end{thebibliography}
\end{document}